\documentclass[10pt,twocolumn,letterpaper]{article}

\usepackage[pagenumbers]{cvpr}
\usepackage[T1]{fontenc}
\usepackage{amsfonts}
\usepackage{amsthm}
\usepackage{algorithm}
\usepackage{algpseudocode}
\usepackage{multirow}
\usepackage{tabularx}

\algrenewcommand\algorithmicrequire{\textbf{Input:}}
\algrenewcommand\algorithmicensure{\textbf{Output:}}

\newcommand{\bestresult}[1]{\textcolor{red}{\textbf{#1}}}
\newcommand{\secondresult}[1]{\textbf{#1}}
\newcommand{\GLOMAPAvg}{GLOMAP-Avg}
\DeclareMathOperator{\argmin}{arg\,min}

\def\BibTeX{{\rm B\kern-.05em{\sc i\kern-.025em b}\kern-.08em
    T\kern-.1667em\lower.7ex\hbox{E}\kern-.125emX}}

\newtheorem{theorem}{Theorem}

\definecolor{cvprblue}{rgb}{0.21,0.49,0.74}
\usepackage[pagebackref,breaklinks,colorlinks,allcolors=cvprblue]{hyperref}
\hypersetup{pdftitle={Towards Rotation-Only Imaging Geometry: Multi-View Pose Estimation}}
\makeatletter
\def\theHALG@line{\thealgorithm.\arabic{ALG@line}}
\makeatother

\begin{document}

\title{Towards Rotation-Only Imaging Geometry:\\ Multi-View 
Pose Estimation}

\hypersetup{pdfauthor={Xinrui Li, Qi Cai, Yuanxin Wu}}
\author{%
Xinrui Li, Qi Cai, and Yuanxin Wu\\[0.3em]
{\normalsize\itshape Shanghai Key Laboratory of Navigation and Location-based Services}\\
{\normalsize\itshape School of Automation and Intelligent Sensing}\\
{\normalsize\itshape Shanghai Jiao Tong University}\\
{\normalsize Shanghai 200240, China}\\
{\normalsize\texttt{physalis@sjtu.edu.cn}, \texttt{qicaicn@gmail.com}, \texttt{yuanx\_wu@hotmail.com}}%
}

\maketitle

\begin{abstract}
Global structure-from-motion (SfM) commonly refines relative poses on individual image pairs and then estimates global rotations by rotation averaging. 
Pairwise refinement does not explicitly combine feature-level constraints with multi-view consistency during rotation initialization. We introduce Global Rotations from Aggregated Matches (GRAM), which estimates global rotations directly from initial relative rotations and inlier matches by minimizing edge-wise rotation-only algebraic residuals. GRAM replaces two-view relative-pose optimization and rotation averaging with a single stage. We further extend rotation-only imaging geometry to multiple views and develop Multi-view Optimization with Rotation-only Reprojection Error (MORE) for nonlinear global rotation refinement. Analytical Rotation-guided Translation (ART) derives locally scaled triplet translations from Linear Global Translation (LiGT) constraints and recovers global translations using weighted least squares. Synthetic and real-world experiments demonstrate the accuracy, robustness, and efficiency of the proposed methods.
\end{abstract}

\section{Introduction}
Recovering camera poses from multiple images is a fundamental task in structure-from-motion (SfM).  Conventional global SfM pipelines first estimate two-view geometry, refine pairwise relative poses, and recover global rotations through rotation averaging~\cite{glomap,OpenMVGPierre}. This two-stage design inherently decouples feature observations from global consistency: pairwise relative refinement is isolated from global constraints, while subsequent rotation averaging operates blindly without revisiting raw feature matches.

Given camera rotations, Linear Global Translation (LiGT)~\cite{QiCai_IJCV,QiCai_TPAMI,cai2024linearrelativeposeestimation} derives constraints from multi-view feature tracks using pose-only imaging geometry, stacks them into a global homogeneous system, and obtains its unit-norm least-squares solution by singular value decomposition (SVD). However, this single null-space formulation inherently couples all feature tracks, making it exceedingly difficult to locally isolate weak or corrupted constraints. While rotation-only imaging geometry~\cite{LXRPAMI} formulates reprojection residuals solely on the rotation manifold, it remains strictly limited to two-view geometry.

Motivated by these insights, we propose a robust global pose-estimation framework based on rotation-only imaging geometry. Our main contributions are:
\begin{itemize}
    \item We introduce Global Rotations from Aggregated Matches (GRAM) to unify pairwise refinement and rotation averaging. By formulating edge-wise residuals directly upon global rotations, GRAM bypasses intermediate steps  and efficiently initializes global rotations in a single stage.
    \item We extend the rotation-only imaging geometry to a multi-view framework.
    By analytically deriving local LiGT-based triplet translations, Multi-view Optimization with Rotation-only Reprojection Error (MORE) further refines the global rotations.
    
    \item We introduce Analytical Rotation-guided Translation (ART), which embeds local LiGT-based triplet translations into a unified weighted linear system to analytically recover all global translations.
\end{itemize}

\section{Related Work}

\textbf{Two-view pose estimation.}
Two-view initialization uses five- to eight-point solvers within RANSAC to estimate epipolar geometry from feature matches~\cite{STEWENIUS2006284,NisterRelative,Pizarro2003RelativePE,MultipleViewGeometry,LONGUETHIGGINS198761,SevenPtHartley}. Sampson-error optimization refines relative pose over inlier matches~\cite{glomap}. Kneip's methods use spherical reprojection or eigenvalue objectives~\cite{KneipOpenGV,Kneip13ICCV}. Linear relative-pose methods eliminate 3D structure~\cite{cai2024linearrelativeposeestimation}. Two-view Reprojection Residuals on the Rotation Manifold (TRRM) enable rotation-only optimization~\cite{LXRPAMI}. All operate on individual image pairs and do not use multi-view feature tracks.

\textbf{Global rotation estimation.}
Global rotation methods either average two-view relative rotations or optimize directly from feature matches. Rotation averaging includes algebraic averaging~\cite{Govindu004}, chordal averaging~\cite{Martinec07}, $L_1$ averaging~\cite{HartleyL1}, iterative reweighting~\cite{ChatterjeePAMI2018}, strong duality~\cite{ErikssonStrongDuality2018}, lifted relaxation~\cite{DellaertShonan2020}, and hybrid optimization~\cite{ChenHybrid2021}. Rotation-Only Bundle Adjustment (ROBA)~\cite{ROBA} utilizes two-view residuals, while Global Reprojection Residual on the Rotation Manifold (GRRM)~\cite{LXRPAMI} aggregates two-view rotation-only reprojection errors. Both methods require traditional two-view estimation and rotation averaging for initialization.

\textbf{Global translation estimation.}
Global translation methods use relative translation directions, feature tracks, or triplet geometry. Direction-based methods improve robustness through filtering and convex estimation~\cite{1dsfm,LUD,BATA2018}. LiGT solves a homogeneous system built from feature tracks~\cite{QiCai_TPAMI}, while Hybrid Explicit Translation Averaging (HETA) and GLOMAP jointly optimize global translations and 3D points~\cite{HETA2024,glomap}. Moulon et al.~\cite{MoulonGlobalFusion2013} estimate triplet translations from trifocal geometry and jointly recover their scales and global translations using constrained $\ell_\infty$ optimization.

\textbf{Joint optimization.}
Bundle adjustment (BA) jointly refines camera poses and 3D points by minimizing reprojection error~\cite{BundleAdjustmentAModernSynthesis}, whereas pose adjustment (PA) eliminates 3D structure and optimizes camera poses alone~\cite{QiCai_TPAMI}. BA commonly serves as the final refinement stage of SfM systems~\cite{COLMAP2016,glomap,OpenMVGPierre,MGSfM2025}.

\section{Mathematical Notation}\label{section:mathematical_notation}
All derivations use normalized camera coordinates. Camera $i$ is represented by
$(\boldsymbol{R}_i,\boldsymbol{t}_i)$, with $\boldsymbol{R}_i$ the world-to-camera rotation and $\boldsymbol{t}_i$ its world-frame center. For the relative pose,
$\boldsymbol{R}_{ij}=\boldsymbol{R}_j\boldsymbol{R}_i^T$ and
$\boldsymbol{t}_{ij}=\boldsymbol{R}_j(\boldsymbol{t}_i-\boldsymbol{t}_j)$.

$|\cdot|$ denotes scalar absolute value or set cardinality,
$\|\boldsymbol{v}\|=\|\boldsymbol{v}\|_2$,
$\hat{\boldsymbol{v}}=\boldsymbol{v}/\|\boldsymbol{v}\|$ for
$\boldsymbol{v}\neq\boldsymbol{0}$, and
$(\boldsymbol{v})_\times\boldsymbol{w}=\boldsymbol{v}\times\boldsymbol{w}$.
$\boldsymbol{M}_{[i,j]}$ and $\boldsymbol{v}_{[i]}$ denote a matrix entry and a vector component, respectively.
$\operatorname{col}(\boldsymbol A_1,\ldots,\boldsymbol A_n)
\triangleq[\boldsymbol A_1^T,\ldots,\boldsymbol A_n^T]^T$ denotes vertical stacking.
$\operatorname{diag}(\boldsymbol A_1,\ldots,\boldsymbol A_n)$ denotes block-diagonal assembly. For indexed blocks $\{\boldsymbol A_i\}_{i=1}^{n}$, $\operatorname{col}_{i=1}^{n}(\boldsymbol A_i)$ and $\operatorname{diag}_{i=1}^{n}(\boldsymbol A_i)$ denote vertical stacking and block-diagonal assembly, respectively.
$[\boldsymbol A_i]_{i=1}^{n}$ denotes horizontal block concatenation.
Let $\boldsymbol{q}(\boldsymbol{R})\in\mathbb{S}^3$ be the unit-quaternion representation of $\boldsymbol{R}$. We define the $SO(3)$ geodesic distance as $d_q(\boldsymbol{R}_i,\boldsymbol{R}_j)=2\cos^{-1}\!\left(\left|\operatorname{Re}\!\left(\boldsymbol{q}(\boldsymbol{R}_i)^{-1}\circ\boldsymbol{q}(\boldsymbol{R}_j)\right)\right|\right)$, where $\circ$ denotes quaternion multiplication and $\operatorname{Re}(\cdot)$ extracts the real part.
For $\boldsymbol B\in\mathbb R^{3\times3}$ with SVD
$\boldsymbol B=\boldsymbol U\boldsymbol\Sigma\boldsymbol V^T$, its nearest
rotation in the Frobenius norm is $\Pi_{SO(3)}(\boldsymbol B)=\boldsymbol U
\operatorname{diag}\!\left(1,1,\det(\boldsymbol U\boldsymbol V^T)\right)
\boldsymbol V^T$. 
For $\boldsymbol A\in\mathbb R^{m\times3}$, $\boldsymbol e_{\min}(\boldsymbol A)$ denotes its analytically computed smallest right singular vector.
For a unique smallest singular value, Cardano's formula gives
$\lambda_{\min}(\boldsymbol A^T\boldsymbol A)$, and all nonzero columns of
$\operatorname{adj}(\boldsymbol A^T\boldsymbol A-\lambda_{\min}\boldsymbol I)$
span the same one-dimensional null space. Normalizing the largest-norm
column yields $\boldsymbol e_{\min}(\boldsymbol A)$, with this choice made
only for numerical stability~\cite{Kneip13ICCV,LXRPAMI}.

Let $\boldsymbol{X}_k$ be a 3D point and
$\boldsymbol{x}_{ik}=(u_{ik},v_{ik},1)^T$ its normalized homogeneous observation in camera $i$. For $\boldsymbol{y}=(y_1,y_2,y_3)^T$ with $y_3\neq0$, define
$\pi(\boldsymbol{y})=(y_1/y_3,y_2/y_3)^T$.
Let $G=(V,E)$ be the view graph and $\mathcal{M}_{ij}$ the feature-match set between cameras $i$ and $j$.
For nonzero coplanar vectors satisfying $\alpha\hat{\boldsymbol{a}}=\beta\hat{\boldsymbol{b}}+\gamma\hat{\boldsymbol{c}}$
with $\alpha,\beta,\gamma>0$, the vector sine rule gives
\begin{equation}\label{eq:sine_rule_coplanar_cross}
\|(\hat{\boldsymbol{b}})_\times\hat{\boldsymbol{c}}\|\hat{\boldsymbol{a}}
=\|(\hat{\boldsymbol{a}})_\times\hat{\boldsymbol{c}}\|\hat{\boldsymbol{b}}
+\|(\hat{\boldsymbol{a}})_\times\hat{\boldsymbol{b}}\|\hat{\boldsymbol{c}}.
\end{equation}

\section{Preliminaries on Rotation-only Imaging Geometry}\label{section: Preliminaries}
\subsection{Pose-only Geometry}\label{section: PPO}
A point $\boldsymbol{X}_k$ observed by camera $i$ satisfies
\begin{equation}\label{eq:projection_function}
z_{ik}\boldsymbol{x}_{ik}=\boldsymbol{R}_i(\boldsymbol{X}_k-\boldsymbol{t}_i),
\end{equation}
where $z_{ik}>0$ is its depth. If cameras $i$ and $j$ observe the same point, expressing both observations in camera $j$ gives
\begin{equation}\label{eq:coplanar_equation}
z_{jk}\boldsymbol{x}_{jk}
=z_{ik}\boldsymbol{R}_{ij}\boldsymbol{x}_{ik}+\boldsymbol{t}_{ij}.
\end{equation}
For a nonzero baseline, positive depths require
$\hat{\boldsymbol{t}}_{ij}$ to follow the chirality-consistent direction of $\boldsymbol{t}_{ij}$. Chirality selection~\cite{QiCai_TPAMI,LXRPAMI} fixes the sign of $\hat{\boldsymbol{t}}_{ij}$. Let $\boldsymbol\theta_k^{ij}=(\boldsymbol R_{ij}\boldsymbol x_{ik})_\times\boldsymbol x_{jk}$, with $\boldsymbol x$ replaced by $\tilde{\boldsymbol x}$ for noisy observations. Applying~\eqref{eq:sine_rule_coplanar_cross} yields the pose-only constraint~\cite{QiCai_IJCV}
\begin{equation}\label{eq:PPO_coord}
\begin{aligned}
&\|(\boldsymbol{R}_{ij}\boldsymbol{x}_{ik})_\times\hat{\boldsymbol{t}}_{ij}\|\boldsymbol{x}_{jk}
=\|(\boldsymbol{x}_{jk})_\times\hat{\boldsymbol{t}}_{ij}\|\boldsymbol{R}_{ij}\boldsymbol{x}_{ik}\\
&{}+\|\boldsymbol\theta_k^{ij}\|\hat{\boldsymbol{t}}_{ij}.
\end{aligned}
\end{equation}
If $\|\boldsymbol{t}_{ij}\|>0$ and
$\|\boldsymbol\theta_k^{ij}\|>0$, the two camera centers and $\boldsymbol{X}_k$ form a nondegenerate chirality-consistent triangle. The analytical depths are
\begin{equation}\label{eq:ana_depth_expression}
z^{ij}_{ik}=\frac{\|\boldsymbol{t}_{ij}\|\|(\boldsymbol{x}_{jk})_\times\hat{\boldsymbol{t}}_{ij}\|}{\|\boldsymbol\theta_k^{ij}\|},\quad
z^{ij}_{jk}=\frac{\|\boldsymbol{t}_{ij}\|\|(\boldsymbol{R}_{ij}\boldsymbol{x}_{ik})_\times\hat{\boldsymbol{t}}_{ij}\|}{\|\boldsymbol\theta_k^{ij}\|}.
\end{equation}
For a track shared by views $i$, $j$, and $l$, $z_{ik}^{ij}=z_{ik}^{il}$ gives
\begin{equation}\label{eq:valid_depth}
z_{jk}^{ij}\boldsymbol{x}_{jk}=z_{ik}^{il}\boldsymbol{R}_{ij}\boldsymbol{x}_{ik}+\boldsymbol{t}_{ij}.
\end{equation}
For a triplet $(i,j,l)$, LiGT~\cite{QiCai_TPAMI} expresses this relation as
\begin{equation}\label{eq:ligt_translation_constraint}
\left(\boldsymbol B_{ijl}^{k}+\boldsymbol C_{ijl}^{k}\right)\boldsymbol t_i
-\boldsymbol B_{ijl}^{k}\boldsymbol t_l
-\boldsymbol C_{ijl}^{k}\boldsymbol t_j=\boldsymbol0.
\end{equation}
where
\begin{equation}\label{eq:ligt_coefficient_matrices}
\begin{gathered}
\boldsymbol B_{ijl}^{k}
  =(\boldsymbol x_{jk})_\times\boldsymbol R_{ij}\boldsymbol x_{ik}
(\boldsymbol\theta_k^{il})^T(\boldsymbol x_{lk})_\times\boldsymbol R_l,\\
\boldsymbol C_{ijl}^{k}
  =\|\boldsymbol\theta_k^{il}\|^2(\boldsymbol x_{jk})_\times\boldsymbol R_j.
\end{gathered}
\end{equation}
\subsection{Two-view Rotation-only Geometry}\label{section: RO}
For $m_{ij}=|\mathcal M_{ij}|$ matches and a given relative rotation $\boldsymbol R_{ij}$,
an observation matrix can be derived as \cite{LXRPAMI,Kneip13ICCV}:
\begin{equation}\label{eq:jointPt_0}
\boldsymbol P_{ij}
=\sum_{k=1}^{m_{ij}}
\boldsymbol\theta_k^{ij}(\boldsymbol\theta_k^{ij})^T,
\end{equation}
At the true rotation in noise-free data, $\boldsymbol P_{ij}\hat{\boldsymbol t}_{ij}=\boldsymbol0$. With noisy matches, $\boldsymbol e_{\min}(\boldsymbol P_{ij})$ analytically gives its unoriented minimum direction, and chirality selection~\cite{QiCai_TPAMI,LXRPAMI} fixes the sign to obtain $\hat{\boldsymbol t}_{ij}^{\boldsymbol R}(\boldsymbol R_{ij})$.
Under noise-free, nondegenerate conditions, substituting $\hat{\boldsymbol{t}}_{ij}^{\boldsymbol{R}}(\boldsymbol R_{ij})$ into~\eqref{eq:PPO_coord} gives the two-view rotation-only constraint~\cite{LXRPAMI}
\begin{equation}\label{eq:constraint_2v}
\begin{aligned}
\boldsymbol{Y}_{jk}^{ij}
&\triangleq\|(\boldsymbol{x}_{jk})_\times\hat{\boldsymbol{t}}_{ij}^{\boldsymbol{R}}\|\boldsymbol{R}_{ij}\boldsymbol{x}_{ik}
+\|\boldsymbol\theta_k^{ij}\|\hat{\boldsymbol{t}}_{ij}^{\boldsymbol{R}}\\
&=\|(\boldsymbol{R}_{ij}\boldsymbol{x}_{ik})_\times
\hat{\boldsymbol{t}}_{ij}^{\boldsymbol{R}}\|\boldsymbol{x}_{jk}.
\end{aligned}
\end{equation}
Thus, $\pi(\boldsymbol{Y}_{jk}^{ij})=\pi(\boldsymbol{x}_{jk})$. With noisy observations, replacing $\boldsymbol{x}$ by $\tilde{\boldsymbol{x}}$ defines the Two-view Reprojection Residuals on the Rotation Manifold (TRRM)
\begin{equation}\label{eq:residual_2v}
\boldsymbol{E}_{jk}^{ij}(\boldsymbol{R}_{ij})
=\pi\!\left(\widetilde{\boldsymbol{Y}}_{jk}^{ij}\right)
-\pi\!\left(\tilde{\boldsymbol{x}}_{jk}\right)\in\mathbb{R}^2.
\end{equation}

\section{Multi-view Rotation-only Imaging Geometry}\label{section:multi_view_rotation_only_geometry}
\subsection{Global Rotation Constraints from Edge Observation Matrices}\label{section:aggregated_match_constraint}
Given global rotations $\{\boldsymbol R_i\}_{i=1}^{n}$, for each matched view pair
$(i,j)$ not classified as \textit{RotationSingular}~\cite{LXRPAMI}, we define the
observation matrix $\boldsymbol P_{ij}(\boldsymbol R_j\boldsymbol R_i^T)$.
Let $\boldsymbol m_{s^*}$ denote the column of its adjugate corresponding to the largest diagonal element. The pairwise constraint residual is
\begin{equation}\label{eq:pair_residual}
r_{ij}(\boldsymbol R_j\boldsymbol R_i^T)
=
\frac{
\left|\det\!\left(\boldsymbol P_{ij}(\boldsymbol R_j\boldsymbol R_i^T)\right)\right|
}{
\|\boldsymbol m_{s^*}\|_2+\varepsilon_P
},
\end{equation}
where \(\varepsilon_P>0\) prevents division by zero for rank-zero scene structures.
With noisy observations, the Binet--Cauchy identity~\cite{LXRPAMI} gives $\operatorname{adj}(\boldsymbol P_{ij})_{[s,s]}=\sum_{u<v}[(\boldsymbol\theta_u^{ij}\times\boldsymbol\theta_v^{ij})_{[s]}]^2$, the accumulated projection energy along axis $s$. Equation~\eqref{eq:pair_residual} therefore uses the adjugate column associated with the maximum-energy axis and evaluates its regularized null-space residual.

\subsection{Rotation-Only Triplet Imaging Geometry}\label{section:three_view_reprojection_geometry}
To extend two-view rotation-only imaging geometry to multiple views, we use LiGT constraints to recover the relative translations.
\begin{theorem}[LiGT-Based Analytical Triplet Translation]\label{thm:ligt_scale_alignment}
For a triplet $\chi=(\eta,\zeta,\nu)$ satisfying the LiGT nondegeneracy conditions~\cite{QiCai_TPAMI}, choose a nonzero \textit{reference-baseline edge} $(\eta,\zeta)$ and fix $\|\boldsymbol t_{\eta\zeta}\|_2=1$. Let $\mathcal K_\chi$ denote the valid tracks shared by all three views. Stacking both baseline orientations gives $\mathcal B_\chi=\operatorname{col}_{k\in\mathcal K_\chi}(\boldsymbol B_{\eta\nu\zeta}^{k},-\boldsymbol B_{\zeta\nu\eta}^{k}-\boldsymbol C_{\zeta\nu\eta}^{k})$ and $\mathcal C_\chi=\operatorname{col}_{k\in\mathcal K_\chi}(\boldsymbol C_{\eta\nu\zeta}^{k},\boldsymbol C_{\zeta\nu\eta}^{k})$. The symmetric weight matrix $\boldsymbol\Omega_L=\operatorname{diag}_{k\in\mathcal K_\chi}(\omega_k\boldsymbol I_3,\omega_k\boldsymbol I_3)$, $\omega_k>0$, makes the recovered triplet geometry invariant to exchanging $\eta$ and $\zeta$. For full-column-rank $\mathcal C_\chi$, let $\boldsymbol T_\chi=(\mathcal C_\chi^T\boldsymbol\Omega_L\mathcal C_\chi)^{-1}\mathcal C_\chi^T\boldsymbol\Omega_L\mathcal B_\chi$. The resulting LiGT-Based analytical triplet translation solution is
\begin{equation}\label{eq:triplet_ligt_translation}
\begin{aligned}
\boldsymbol S_\chi
&=\mathcal B_\chi^{T}\boldsymbol\Omega_L
(\mathcal B_\chi-\mathcal C_\chi\boldsymbol T_\chi),\\
\boldsymbol t_{\eta\zeta}^{\boldsymbol R,L}
&=\boldsymbol R_\zeta\boldsymbol e_{\min}(\boldsymbol S_\chi),\\
\boldsymbol t_{\eta\nu}^{\boldsymbol R,L}
&=-\boldsymbol R_\nu\boldsymbol T_\chi
\boldsymbol e_{\min}(\boldsymbol S_\chi),\\
\boldsymbol t_{\zeta\nu}^{\boldsymbol R,L}
&=-\boldsymbol R_\nu(\boldsymbol T_\chi+\boldsymbol I_3)
\boldsymbol e_{\min}(\boldsymbol S_\chi).
\end{aligned}
\end{equation}
With a unique minimum eigendirection of $\boldsymbol S_\chi$,~\eqref{eq:triplet_ligt_translation} gives the weighted least-squares solution up to a common sign.
\end{theorem}
\begin{proof}[Proof]
Using $\boldsymbol b_{\eta p}=\boldsymbol t_\eta-\boldsymbol t_p$,
Eq.~\eqref{eq:ligt_translation_constraint} becomes
\begin{equation}\label{eq:supp-art-ligt-baseline-constraint}
\boldsymbol B_{\eta\nu\zeta}^{k}\boldsymbol b_{\eta\zeta}
+\boldsymbol C_{\eta\nu\zeta}^{k}\boldsymbol b_{\eta\nu}
=\boldsymbol0.
\end{equation}
Applying
Eq.~\eqref{eq:supp-art-ligt-baseline-constraint} to the two baseline
orientations $(\eta,\nu,\zeta)$ and $(\zeta,\nu,\eta)$ gives
\begin{equation}\label{eq:supp-art-ligt-stacked-constraint}
\begin{bmatrix}
\boldsymbol B_{\eta\nu\zeta}^{k} &
\boldsymbol C_{\eta\nu\zeta}^{k}\\
-\boldsymbol B_{\zeta\nu\eta}^{k}
-\boldsymbol C_{\zeta\nu\eta}^{k} &
\boldsymbol C_{\zeta\nu\eta}^{k}
\end{bmatrix}
\begin{bmatrix}
\boldsymbol b_{\eta\zeta}\\
\boldsymbol b_{\eta\nu}
\end{bmatrix}
=\boldsymbol0.
\end{equation}
The second row follows from
$\boldsymbol b_{\zeta\eta}=-\boldsymbol b_{\eta\zeta}$ and
$\boldsymbol b_{\zeta\nu}
=\boldsymbol b_{\eta\nu}-\boldsymbol b_{\eta\zeta}$.
Stacking Eq.~\eqref{eq:supp-art-ligt-stacked-constraint} over
$k\in\mathcal K_\chi$ yields
$\mathcal B_\chi\boldsymbol b_{\eta\zeta}
+\mathcal C_\chi\boldsymbol b_{\eta\nu}=\boldsymbol0$,
where
\begin{equation}\label{eq:supp-art-ligt-double-baseline-matrices}
\begin{gathered}
\mathcal B_\chi
=\operatorname{col}_{k\in\mathcal K_\chi}\!\left(
\boldsymbol B_{\eta\nu\zeta}^{k},
-\boldsymbol B_{\zeta\nu\eta}^{k}
-\boldsymbol C_{\zeta\nu\eta}^{k}\right),\\
\mathcal C_\chi
=\operatorname{col}_{k\in\mathcal K_\chi}\!\left(
\boldsymbol C_{\eta\nu\zeta}^{k},
\boldsymbol C_{\zeta\nu\eta}^{k}\right).
\end{gathered}
\end{equation}
Let
$\boldsymbol\Omega_L
=\operatorname{diag}_{k\in\mathcal K_\chi}
(\omega_k\boldsymbol I_3,\omega_k\boldsymbol I_3)$ with
$\omega_k>0$. Under $\|\boldsymbol b_{\eta\zeta}\|_2=1$, we obtain the
following weighted least-squares problem:
\begin{equation}\label{eq:supp-art-ligt-vector-objective}
\underset{\|\boldsymbol b_{\eta\zeta}\|_2=1,\,
\boldsymbol b_{\eta\nu}}{\min}\quad
\left\|\boldsymbol\Omega_L^{1/2}
(\mathcal B_\chi\boldsymbol b_{\eta\zeta}
+\mathcal C_\chi\boldsymbol b_{\eta\nu})
\right\|_2^2.
\end{equation}
For full-column-rank $\mathcal C_\chi$, define
\begin{equation}\label{eq:supp-art-ligt-eliminated-translation}
\boldsymbol T_\chi
=(\mathcal C_\chi^{T}\boldsymbol\Omega_L\mathcal C_\chi)^{-1}
\mathcal C_\chi^{T}\boldsymbol\Omega_L\mathcal B_\chi.
\end{equation}
For fixed $\boldsymbol b_{\eta\zeta}$, minimization over
$\boldsymbol b_{\eta\nu}$ gives
$\boldsymbol b_{\eta\nu}=-\boldsymbol T_\chi\boldsymbol b_{\eta\zeta}$.
Substituting it into Eq.~\eqref{eq:supp-art-ligt-vector-objective}
eliminates $\boldsymbol b_{\eta\nu}$:
\[
\left\|\boldsymbol\Omega_L^{1/2}
(\mathcal B_\chi-\mathcal C_\chi\boldsymbol T_\chi)
\boldsymbol b_{\eta\zeta}\right\|_2^2
=\boldsymbol b_{\eta\zeta}^T\boldsymbol S_\chi
\boldsymbol b_{\eta\zeta},
\]
where
\begin{equation}\label{eq:supp-art-ligt-schur-matrix}
\begin{aligned}
\boldsymbol S_\chi
&=\mathcal B_\chi^{T}\boldsymbol\Omega_L\mathcal B_\chi
-\mathcal B_\chi^{T}\boldsymbol\Omega_L\mathcal C_\chi
(\mathcal C_\chi^{T}\boldsymbol\Omega_L\mathcal C_\chi)^{-1}
\mathcal C_\chi^{T}\boldsymbol\Omega_L\mathcal B_\chi\\
&=\mathcal B_\chi^{T}\boldsymbol\Omega_L
(\mathcal B_\chi-\mathcal C_\chi\boldsymbol T_\chi).
\end{aligned}
\end{equation}
Together with Eq.~\eqref{eq:supp-art-ligt-schur-matrix},
Eq.~\eqref{eq:supp-art-ligt-vector-objective} is reduced to
\[
\min_{\|\boldsymbol b_{\eta\zeta}\|_2=1}
\boldsymbol b_{\eta\zeta}^T\boldsymbol S_\chi
\boldsymbol b_{\eta\zeta}.
\]
By the Rayleigh--Ritz theorem,
$\boldsymbol b_{\eta\zeta}=\boldsymbol e_{\min}(\boldsymbol S_\chi)$.
Together with the eliminated relation
$\boldsymbol b_{\eta\nu}=-\boldsymbol T_\chi\boldsymbol b_{\eta\zeta}$
and the closure
$\boldsymbol b_{\zeta\nu}
=\boldsymbol b_{\eta\nu}-\boldsymbol b_{\eta\zeta}$, this gives
\[
\boldsymbol b_{\eta\nu}
=-\boldsymbol T_\chi\boldsymbol e_{\min}(\boldsymbol S_\chi),
\qquad
\boldsymbol b_{\zeta\nu}
=-(\boldsymbol T_\chi+\boldsymbol I_3)
\boldsymbol e_{\min}(\boldsymbol S_\chi).
\]
The corresponding relative translations in the camera coordinate systems are
\begin{equation}\label{eq:supp-art-ligt-triplet-solution}
\begin{aligned}
\boldsymbol t_{\eta\zeta}^{\boldsymbol R,L}
&=\boldsymbol R_\zeta\boldsymbol b_{\eta\zeta}
=\boldsymbol R_\zeta\boldsymbol e_{\min}(\boldsymbol S_\chi),\\
\boldsymbol t_{\eta\nu}^{\boldsymbol R,L}
&=\boldsymbol R_\nu\boldsymbol b_{\eta\nu}
=-\boldsymbol R_\nu\boldsymbol T_\chi
\boldsymbol e_{\min}(\boldsymbol S_\chi),\\
\boldsymbol t_{\zeta\nu}^{\boldsymbol R,L}
&=\boldsymbol R_\nu\boldsymbol b_{\zeta\nu}
=\boldsymbol R_\nu(\boldsymbol b_{\eta\nu}-\boldsymbol b_{\eta\zeta})\\
&=-\boldsymbol R_\nu(\boldsymbol T_\chi+\boldsymbol I_3)
\boldsymbol e_{\min}(\boldsymbol S_\chi).
\end{aligned}
\end{equation}
Since the two baseline orientations of each track share the same weight,
exchanging $\eta$ and $\zeta$ transforms
$(\boldsymbol b_{\eta\zeta},\boldsymbol b_{\eta\nu})$ into
$(-\boldsymbol b_{\eta\zeta},
\boldsymbol b_{\eta\nu}-\boldsymbol b_{\eta\zeta})$
and only permutes the two residual blocks in
Eq.~\eqref{eq:supp-art-ligt-stacked-constraint}. Hence, the LiGT-Based
analytical triplet translation solution is invariant to choosing
$(\eta,\zeta,\nu)$ or $(\zeta,\eta,\nu)$ as the ordered triplet. For
full-column-rank $\mathcal C_\chi$ and a simple smallest eigenvalue of
$\boldsymbol S_\chi$, the solution is unique up to a common sign.
Equations~\eqref{eq:supp-art-ligt-stacked-constraint}--%
\eqref{eq:supp-art-ligt-triplet-solution} complete the proof of Theorem~1.
\end{proof}

If all three camera centers coincide, no nonzero \textit{reference-baseline edge} exists and
$\mathcal C_\chi=\boldsymbol0$. Hence,
$\operatorname{rank}(\mathcal C_\chi)=0$, and the triplet is excluded by the
full-column-rank condition.

Under noise, the choice of \textit{reference-baseline edge} affects
triplet-translation accuracy. We select it based on eigenvalue conditioning to avoid degeneracy.

For the selected \textit{reference-baseline edge}
$(i,l)$, Theorem~\ref{thm:ligt_scale_alignment} gives $\boldsymbol t_{il}^{\boldsymbol R,L}$
and $\boldsymbol t_{ij}^{\boldsymbol R,L}$ with a common local scale.
After resolving the common sign using the chirality constraint~\cite{QiCai_TPAMI}, we use the analytical depth
in~\eqref{eq:ana_depth_expression} to write~\eqref{eq:valid_depth}
as the rotation-only triplet constraint
\begin{equation}\label{eq:MORE_constraint}
\begin{gathered}
\boldsymbol{Y}_{jik}^{il}
\triangleq\|(\boldsymbol{t}_{il}^{\boldsymbol R,L})_\times
\boldsymbol{x}_{lk}\|\boldsymbol{R}_{ij}\boldsymbol{x}_{ik}\\
{}+\|(\boldsymbol{x}_{lk})_\times\boldsymbol{R}_{il}
\boldsymbol{x}_{ik}\|\boldsymbol{t}_{ij}^{\boldsymbol R,L},\\
\pi(\boldsymbol{Y}_{jik}^{il})=\pi(\boldsymbol{x}_{jk}).
\end{gathered}
\end{equation}
For noisy observations, replacing $\boldsymbol x$ by
$\tilde{\boldsymbol x}$ gives
$\widetilde{\boldsymbol Y}_{jik}^{il}$ and the residual
\begin{equation}\label{eq:MORE_residual}
\boldsymbol{\epsilon}_{jik}^{il}
=\pi(\widetilde{\boldsymbol{Y}}_{jik}^{il})
-\pi(\tilde{\boldsymbol{x}}_{jk})\in\mathbb{R}^2.
\end{equation}
Here, $j$ is the target view, $(i,l)$ is the \textit{reference-view pair}, and $k$ indexes the feature track.
Since $\pi(\alpha\widetilde{\boldsymbol{Y}})=\pi(\widetilde{\boldsymbol{Y}})$, the
residual is invariant to a common nonzero translation scale
and reduces to the TRRM residual in~\eqref{eq:residual_2v} when $j=l$.
For the selected tuples $\mathcal S$
and robust loss $\rho$, Multi-view Optimization with Rotation-only Reprojection Error (MORE) solves
\begin{equation}\label{eq:more_objective}
\{\boldsymbol R_i^*\}_{i=1}^{n}
=\underset{\substack{\boldsymbol R_i\in SO(3)\\
\boldsymbol R_a=\boldsymbol I}}{\arg\min}
\sum_{(j,k,i,l)\in\mathcal S}
\rho\!\left(
\|\boldsymbol\epsilon_{jik}^{il}\|_2^2
+\|\boldsymbol\epsilon_{jlk}^{li}\|_2^2
\right).
\end{equation}

\subsection{Analytical Rotation-guided Global Translation}\label{section:global_translation}
Theorem~\ref{thm:ligt_scale_alignment} analytically recovers triplet translations from global rotations and feature observations. We then follow the per-triplet registration formulation of Moulon et al.~\cite{MoulonGlobalFusion2013} to recover the global translations analytically.

Let $\mathcal T=\{\chi_p=(i_p,l_p,j_p)\}_{p=1}^{h}$ contain
$h$ valid triplets covering a connected camera graph, with
$(i_p,l_p)$ the selected \textit{reference-baseline edge} of
$\chi_p$. Choose a reference camera $r$, set $\boldsymbol t_r=\boldsymbol0$, and stack the translations to be estimated as $\boldsymbol t=\operatorname{col}_{i\ne r}(\boldsymbol t_i)\in\mathbb R^{3(n-1)}$. For each $\chi_p$ with coefficient \(\alpha_p\in\mathbb R\), Theorem~\ref{thm:ligt_scale_alignment} gives
\begin{equation}\label{eq:art_triplet_constraint}
\boldsymbol q_p=\operatorname{col}\!\left(
\boldsymbol R_{l_p}\boldsymbol e_{\min}(\boldsymbol S_{\chi_p}),
-\boldsymbol R_{j_p}\boldsymbol T_{\chi_p}
\boldsymbol e_{\min}(\boldsymbol S_{\chi_p})\right).
\end{equation}
Let
$\boldsymbol E_{p,i_p}=\operatorname{col}(\boldsymbol R_{l_p},\boldsymbol R_{j_p})$,
$\boldsymbol E_{p,l_p}=\operatorname{col}(-\boldsymbol R_{l_p},\boldsymbol0)$, and
$\boldsymbol E_{p,j_p}=\operatorname{col}(\boldsymbol0,-\boldsymbol R_{j_p})$.
$\boldsymbol E_{p,k}=\boldsymbol0$ when
$k\notin\{i_p,l_p,j_p\}$. With
$\boldsymbol E_p=[\boldsymbol E_{p,k}]_{k=1,\,k\ne r}^{n}$, the constraint is
$\boldsymbol E_p\boldsymbol t=\alpha_p\boldsymbol q_p$.
Let $\boldsymbol E=\operatorname{col}_{p=1}^{h}(\boldsymbol E_p)$ and $\boldsymbol Q=\operatorname{diag}_{p=1}^{h}(\boldsymbol q_p)$. Setting $\alpha_1=1$, write $\boldsymbol Q=[\boldsymbol Q_1\;\boldsymbol Q_2]$ and $\boldsymbol\alpha_2=\operatorname{col}_{p=2}^{h}(\alpha_p)$. Define
\begin{equation}\label{eq:art_augmented_system}
\boldsymbol H=\left[\boldsymbol E\;-\boldsymbol Q_2\right],
\quad \boldsymbol d=\boldsymbol Q_1,\quad
\boldsymbol x=\operatorname{col}\!\left(\boldsymbol t,\boldsymbol\alpha_2\right).
\end{equation}
Let $\boldsymbol\Omega_A\in\mathbb R^{6h\times6h}$ be the positive-definite weight matrix. For full-column-rank $\boldsymbol H$, the solution is
\begin{equation}\label{eq:art_closed_form}
\begin{gathered}
\boldsymbol x^*=(\boldsymbol H^T\boldsymbol\Omega_A\boldsymbol H)^{-1}
\boldsymbol H^T\boldsymbol\Omega_A\boldsymbol d,\\
\boldsymbol t_{\rm ART}
=\left[\boldsymbol I_{3(n-1)}\;
\boldsymbol 0_{3(n-1)\times(h-1)}\right]\boldsymbol x^*,
\qquad \boldsymbol t_r^*=\boldsymbol0.
\end{gathered}
\end{equation}
Equations~\eqref{eq:triplet_ligt_translation} and~\eqref{eq:art_triplet_constraint}--\eqref{eq:art_closed_form} define Analytical Rotation-guided Translation (ART). For fixed $\boldsymbol\Omega_A$, global translations are recovered analytically from global rotations and feature observations. Finally, the sign of $\boldsymbol t_{\mathrm{ART}}$ is resolved using the chirality constraint~\cite{QiCai_TPAMI}.
In contrast, Moulon et al.~\cite{MoulonGlobalFusion2013} estimate triplet translations through trifocal geometry and local BA, then register them globally using constrained \(\ell_\infty\) optimization.

\section{Robust Global Pose Estimation}\label{section:robust_global_pose_estimation}

\subsection{Global Rotation Initialization}
GRAM initializes global rotations in three stages. First, edges satisfying $c_{ij}=\operatorname*{median}_{l}d_q(\boldsymbol R_{li}\boldsymbol R_{jl}\boldsymbol R_{ij},\boldsymbol I)<\tau_g$ form the reliable edge set $\mathcal E_g$, dividing the view graph into initial subgraphs. Second, within each subgraph $\mathcal C$ (with reliable edges $\mathcal E_g^{\mathcal C}$ and an anchored view $a\in\mathcal C$), rotations are initialized via a $c_{ij}$-weighted minimum spanning tree (MST). They are then jointly refined by solving:
\begin{equation}\label{eq:gram_subgraph_refinement}
\{\boldsymbol R_i^{\mathcal C,*}\}_{i\in\mathcal C}
=\underset{\substack{\boldsymbol R_i^{\mathcal C}\in SO(3)\\ \boldsymbol R_a^{\mathcal C}=\boldsymbol I}}{\arg\min}
\sum_{(i,j)\in\mathcal E_g^{\mathcal C}}
r_{ij}\!\left(\boldsymbol R_j^{\mathcal C}(\boldsymbol R_i^{\mathcal C})^T\right)^2 .
\end{equation}
Third, we incrementally merge adjacent subgraphs $\mathcal A$ and $\mathcal B$. Each cross-edge $(i,j)$ with $i\in\mathcal A$ and $j\in\mathcal B$ yields an alignment candidate $\boldsymbol Q_{\mathcal B\to\mathcal A}^{(ij)}=(\boldsymbol R_j^{\mathcal B})^T\boldsymbol R_{ij}\boldsymbol R_i^{\mathcal A}$. The largest $d_q$-cluster among them forms the reliable cross-edge set $\mathcal I_{\mathcal A\mathcal B}$. We project their mean onto $SO(3)$ to initialize the shared alignment $\boldsymbol Q_{\mathcal B\to\mathcal A}$, and further refine it via:
\begin{equation}\label{eq:gram_subgraph_alignment}
\boldsymbol Q_{\mathcal B\to\mathcal A}^{*}
=\underset{\boldsymbol Q\in SO(3)}{\arg\min}
\sum_{(i,j)\in\mathcal I_{\mathcal A\mathcal B}}
r_{ij}\!\left(
\boldsymbol R_j^{\mathcal B}\boldsymbol Q
(\boldsymbol R_i^{\mathcal A})^T
\right)^2 .
\end{equation}
This process repeats until all subgraphs are merged, as summarized in Algorithm~\ref{alg:gram}.

\begin{algorithm}[H]
\caption{GRAM: Global Rotations from Aggregated Matches}
\label{alg:gram}
\footnotesize
\begin{algorithmic}[1]
\Require View graph $G=(V,E)$, relative rotations $\{\boldsymbol R_{ij}\}_{(i,j)\in E}$, inlier matches $\{\mathcal M_{ij}\}_{(i,j)\in E}$, and thresholds $\tau_g,\tau_c, \tau_r$
\Ensure Initial global rotations $\{\boldsymbol R_i^{(0)}\}_{i\in V}$ and reliable edge set $\mathcal E_{\mathrm{reg}}$
\State $\mathcal E_{\mathrm{reg}}\gets\{(i,j)\in E:c_{ij}<\tau_g\}$; form initial subgraphs
\State Per subgraph, use $c_{ij}$-MST $\mathcal E_{\mathrm{MST}}$; apply~\eqref{eq:gram_subgraph_refinement} if $\max_{(i,j)\in\mathcal E_{\mathrm{MST}}}c_{ij}>\tau_r$
\State Let $\mathcal A$ be the initial subgraph with the most views
\While{an unmerged subgraph $\mathcal B$ is adjacent to $\mathcal A$}
    \State Identify reliable cross-edges $\mathcal I_{\mathcal A\mathcal B}$ via candidate clustering
    \State Align $\mathcal B$ to $\mathcal A$ by~\eqref{eq:gram_subgraph_alignment}; $\mathcal E_{\mathrm{reg}}\gets\mathcal E_{\mathrm{reg}}\cup\mathcal I_{\mathcal A\mathcal B}$; merge $\mathcal B$ into $\mathcal A$
\EndWhile
\State \Return Assembled $\{\boldsymbol R_i^{(0)}\}_{i\in V}$ and $\mathcal E_{\mathrm{reg}}$
\end{algorithmic}
\end{algorithm}

\subsection{Global Rotation Refinement}
\begin{algorithm}[H]
\caption{MORE: Multi-view Optimization with
Rotation-only Reprojection Error}
\label{alg:more}
\footnotesize
\begin{algorithmic}[1]
\Require Initial rotations $\{\boldsymbol R_i^{(0)}\}_{i=1}^{n}$, reliable edges, and inlier matches
\Ensure Refined global rotations $\{\boldsymbol R_i^*\}_{i=1}^{n}$
\State Recover chirality-consistent local translations by Theorem~\ref{thm:ligt_scale_alignment}
\State For each observation $\tilde{\boldsymbol x}_{jk}$, select the valid three-view reference-view pair $(i,l)$ maximizing $\|\boldsymbol\theta_k^{il}\|$, or use the valid two-view case $(i,j)$ by setting $l=j$, and add $(j,k,i,l)$ to $\mathcal S$.
\State Optimize~\eqref{eq:more_objective} from $\{\boldsymbol R_i^{(0)}\}$
\end{algorithmic}
\end{algorithm}

\subsection{Analytical Global Translation Estimation}
As summarized in Algorithm~\ref{alg:art}, ART performs robust triplet screening and IRLS weighting, with each weighted subproblem solved analytically. Parameter details are deferred to the supplement.

\begin{algorithm}[H]
\caption{ART: Analytical Rotation-guided Translation}
\label{alg:art}
\footnotesize
\begin{algorithmic}[1]
\Require Global rotations $\{\boldsymbol R_i\}_{i=1}^{n}$, candidate triplets $\mathcal T$, and inlier tracks
\Ensure Global translations $\boldsymbol t_{\mathrm{ART}}$
\State Recover locally scaled triplet translations by Theorem~\ref{thm:ligt_scale_alignment}
\State Filter observations, recompute translations, and discard degenerate triplets
\State $\mathcal T^{\mathrm{ret}}\gets$ Select triplets by robust confidence while preserving connectivity
\State Update $\boldsymbol\Omega_A$ by IRLS and evaluate~\eqref{eq:art_closed_form} analytically after gauge fixing
\State Resolve the common sign of $\boldsymbol t_{\mathrm{ART}}$ using the chirality constraint
\State \Return $\boldsymbol t_{\mathrm{ART}}$
\end{algorithmic}
\end{algorithm}

\section{Experiments}
\subsection{Experimental Setup}

\textbf{Hardware.}
All experiments were run on Ubuntu 22.04 with an Intel Core i9-13900K CPU and 64~GB RAM. GLOMAP and our pipeline use one CPU core and single-threaded solvers.

\begin{figure}[!t]
    \centering
    \includegraphics[width=0.83\columnwidth]{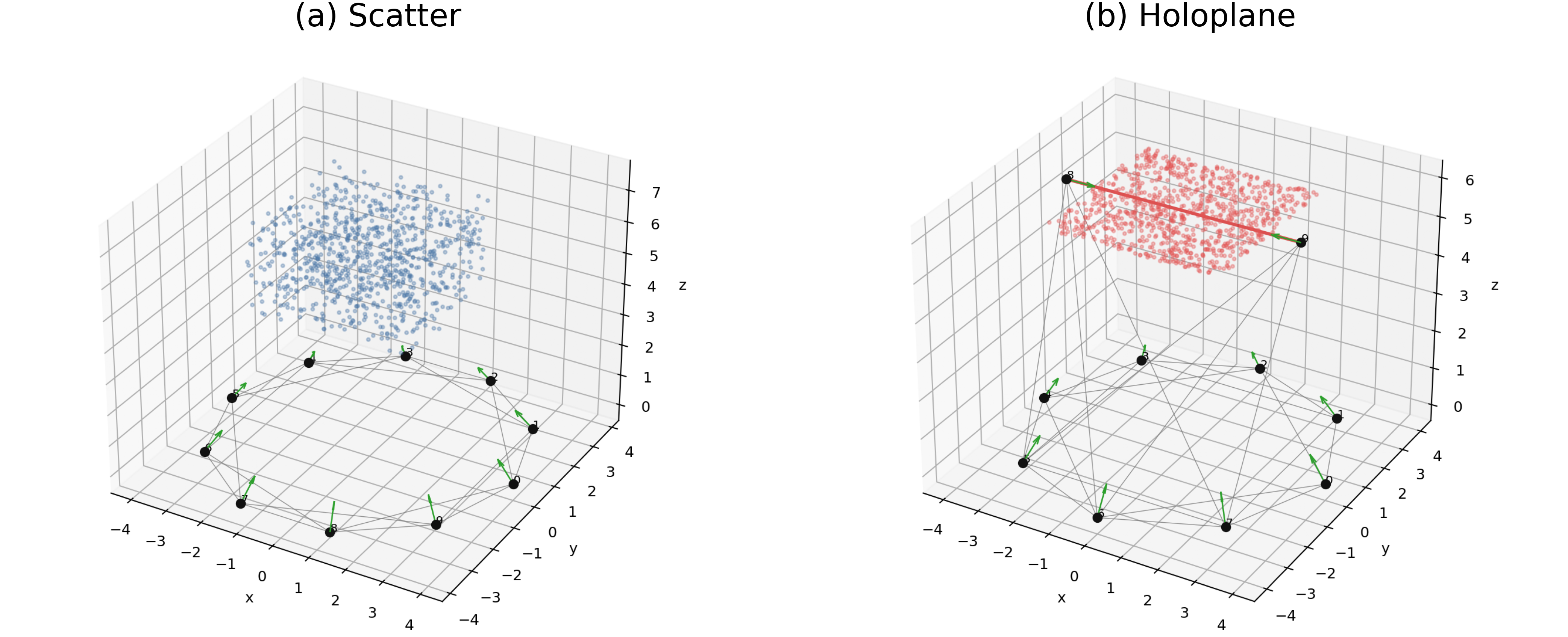}
    \caption{Synthetic 3D point configurations: spatial points and holoplane degeneracy (all ﬁnite 3D points lie on the same plane as the corresponding cameras)~\cite{LXRPAMI}. }
    \label{fig:simulation_scene_configurations}
\end{figure}

\textbf{Synthetic data.}
Following OpenGV~\cite{KneipOpenGV}, we simulate 10 cameras within a unit sphere, with rotation components uniformly sampled from $[-0.5,0.5]$ rad. The complete 45-edge view graph is generated from 1000 3D points shared across all views. We use images of size $2160\times2160$ with $f=800$ px and add uniform noise $\mathcal U[-\sigma,\sigma]$ px for $\sigma\in\{0.1,0.5,1,2,3,4,5,10\}$, with 500 Monte Carlo runs per level. We evaluate spatial scatter, local pure rotation, one corrupted matching edge, and holoplane degeneracy~\cite{LXRPAMI}, as shown in Fig.~\ref{fig:simulation_scene_configurations}.

\textbf{Real data.} We use six Strecha scenes~\cite{StrechaDataset} and 13 ETH3D scenes~\cite{ETH3D2017}, all with accurate camera-pose ground truth.

\textbf{Our pipeline.}
Using COLMAP~\cite{COLMAP2016} camera models and matches and GLOMAP~\cite{glomap} view-graph calibration, PoseLib's Nist\'er five-point RANSAC~\cite{NisterRelative} estimates initial relative poses and inliers without two-view nonlinear refinement. GRAM, MORE, and ART then recover global poses without global BA.

\textbf{Compared methods.}
We mainly compare with GLOMAP~\cite{glomap} and its core modules. The standard LiGT formulation~\cite{QiCai_TPAMI} is also included in the fixed-rotation global-translation comparison and is evaluated using the same verified tracks. All methods use the same fixed intrinsics. Additional baselines are reported in the supplementary material.

\textbf{Metrics.}
Superscripts $\mathrm{est}$ and $\mathrm{gt}$ distinguish estimates from ground truth for global rotations $\boldsymbol R_i$ and translations $\boldsymbol t_i$. Using the $SO(3)$ geodesic distance $d_q$ from Section~\ref{section:mathematical_notation}, the median global-rotation error in degrees is
\begin{equation}
\begin{aligned}
&\boldsymbol Q_R^\star
=\argmin_{\boldsymbol Q\in SO(3)}
\sum_{i=1}^{n}
d_q\!\left(\boldsymbol R_i^{\mathrm{est}}\boldsymbol Q,
\boldsymbol R_i^{\mathrm{gt}}\right)^2,\\
&e_R
=\underset{1\leq i\leq n}{\operatorname{median}}
\frac{180}{\pi}\,
d_q\!\left(\boldsymbol R_i^{\mathrm{est}}\boldsymbol Q_R^\star,
\boldsymbol R_i^{\mathrm{gt}}\right).
\end{aligned}
\label{eq:evaluation_rotation_error}
\end{equation}
With $s^\star>0$, $\boldsymbol Q_t^\star\in SO(3)$, and $\boldsymbol c^\star\in\mathbb R^3$ denoting the least-squares 7-DoF alignment to ground truth, the median global-translation error is
\begin{equation}
e_t
=\underset{1\leq i\leq n}{\operatorname{median}}
\left\|s^\star\boldsymbol Q_t^\star\boldsymbol t_i^{\mathrm{est}}
+\boldsymbol c^\star-\boldsymbol t_i^{\mathrm{gt}}\right\|_2.
\label{eq:evaluation_translation_error}
\end{equation}

\subsection{Synthetic Experiments}

\subsubsection{Two-view Relative Rotation}
\begin{figure}[!t]
    \centering
    \includegraphics[width=0.90\columnwidth]{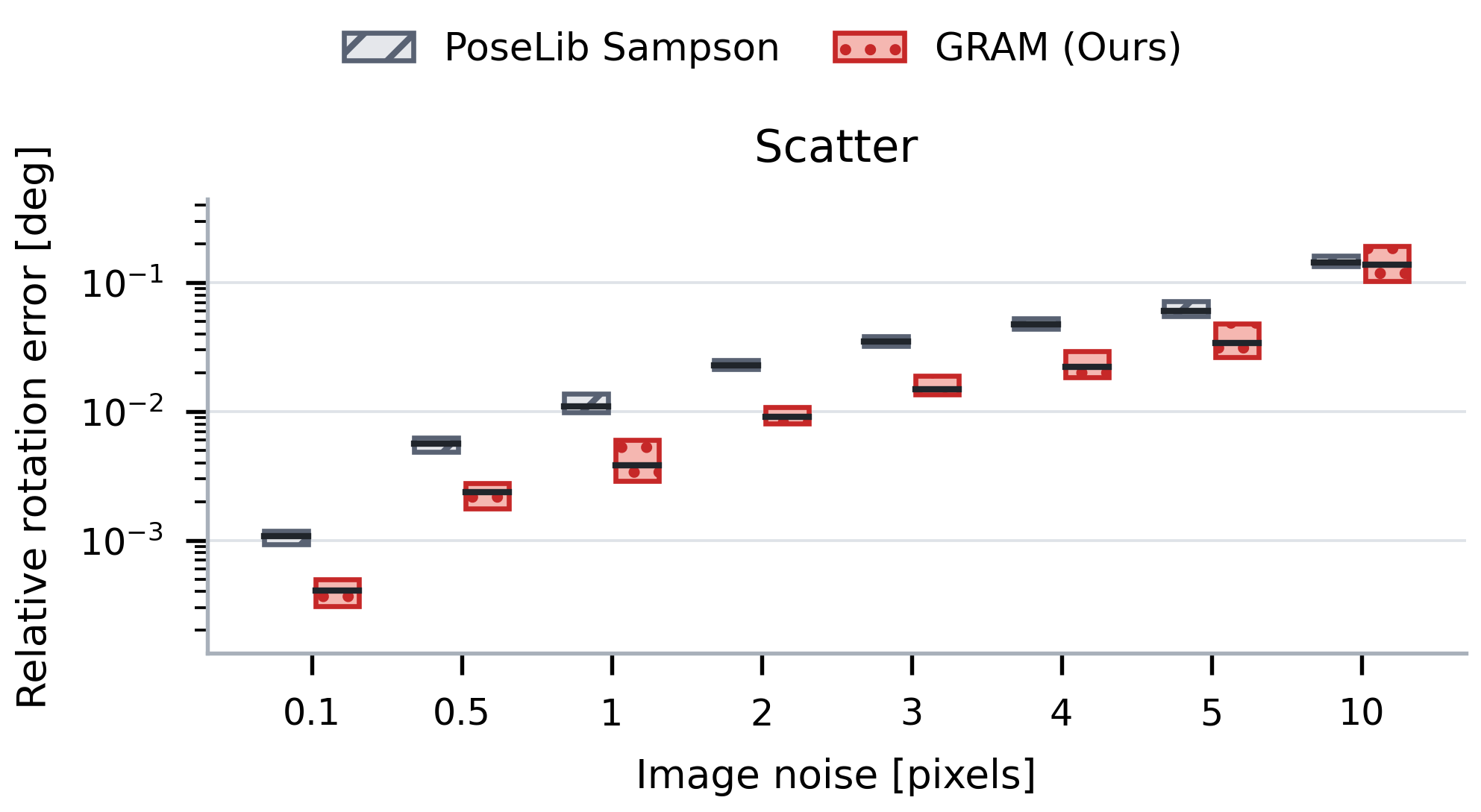}
    \caption{Two-view relative-rotation errors for PoseLib Sampson~\cite{PoseLib} and GRAM (IQRs and medians, log scale).}
    \label{fig:simulation_scatter_two_view}
\end{figure}

In the spatial-scatter setting, we compare GRAM with PoseLib Sampson, which applies five-point RANSAC followed by iterative Sampson-error relative-pose refinement. GRAM is evaluated using the pairwise relative rotations induced by its global rotations. Each boxplot in Fig.~\ref{fig:simulation_scatter_two_view} summarizes 500 run-wise median errors over 45 image pairs. GRAM achieves lower median errors across all noise levels, with roughly a $60\%$ reduction for $\sigma\leq1$ px; the gap narrows as noise increases. Further comparisons~\cite{BundleAdjustmentAModernSynthesis,QiCai_TPAMI,KneipECCV2012,KneipOpenGV,LXRPAMI} in the supplementary material show the same overall trend against additional baselines.

\begin{figure*}[!t]
    \centering
    \captionsetup[subfigure]{font=footnotesize,skip=0pt}
    \begin{subfigure}[t]{0.93\textwidth}
        \centering
        \caption{Global rotation initialization.}
        \label{fig:verified_global_rotation_compact}
        \includegraphics[width=\linewidth]{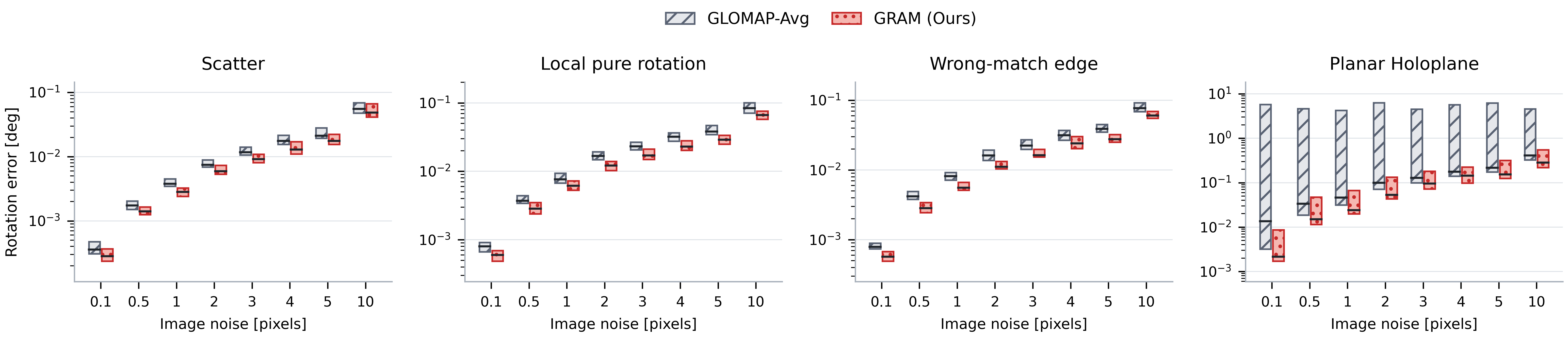}
    \end{subfigure}\vspace{-2pt}

    \begin{subfigure}[t]{0.93\textwidth}
        \centering
        \caption{Global translation with fixed \GLOMAPAvg{} rotations~\cite{glomap}.}
        \label{fig:verified_translation_compact}
        \includegraphics[width=\linewidth]{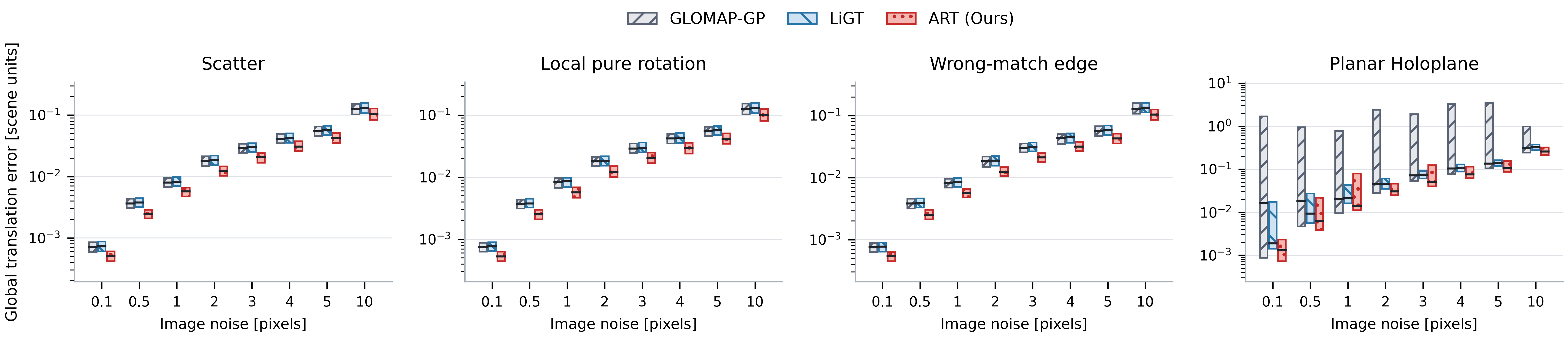}
    \end{subfigure}\vspace{-2pt}

    \begin{subfigure}[t]{0.93\textwidth}
        \centering
        \caption{Global-rotation error in complete pipelines.}
        \label{fig:verified_full_rotation_compact}
        \includegraphics[width=\linewidth]{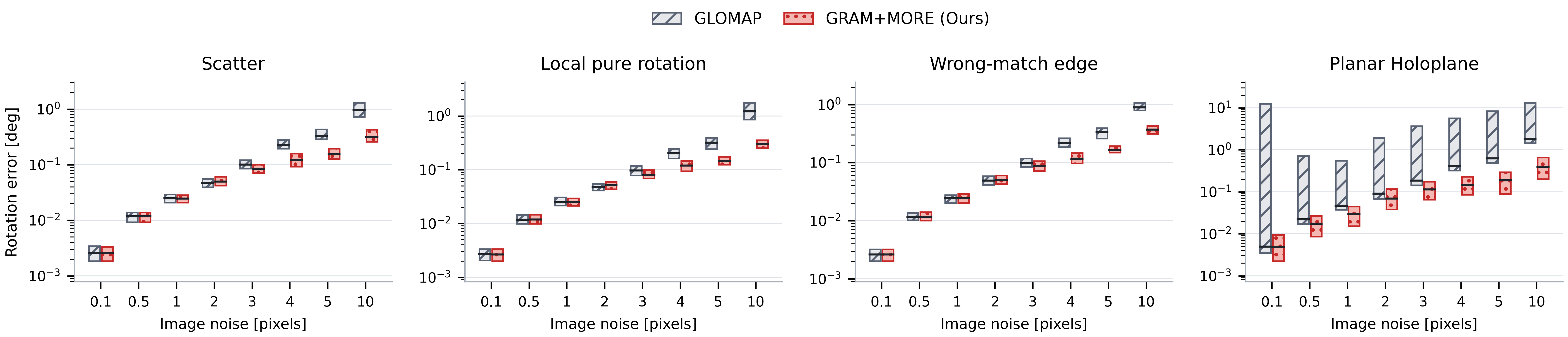}
    \end{subfigure}\vspace{-2pt}

    \begin{subfigure}[t]{0.93\textwidth}
        \centering
        \caption{Global-translation error in complete pipelines.}
        \label{fig:verified_art_after_more_center}
        \includegraphics[width=\linewidth]{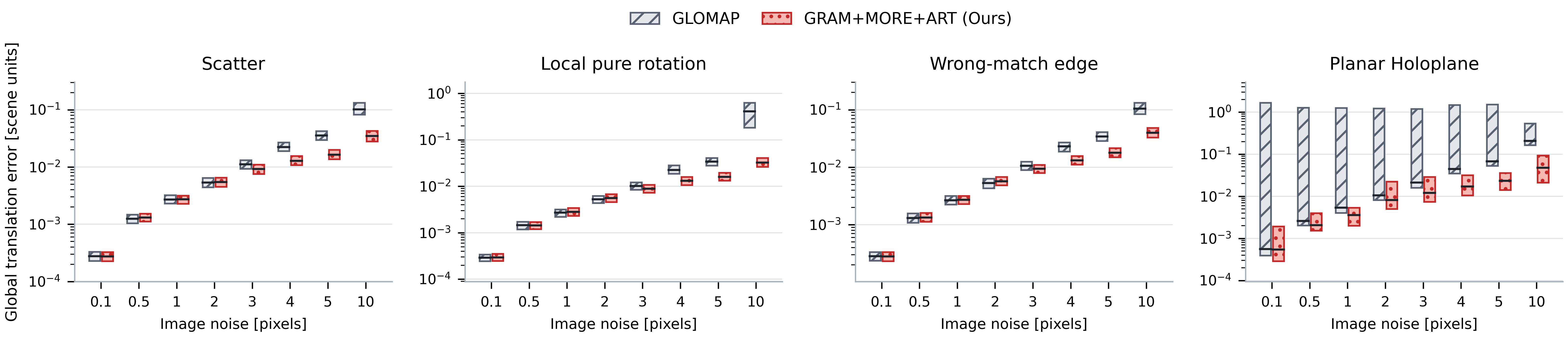}
    \end{subfigure}
    \caption{Four global synthetic experiments. Boxes and black lines show IQRs and medians.}
    \label{fig:global_simulation_two_method_comparison}
\end{figure*}
\begin{table*}[!t]
\caption{Median errors for three experiments on real-world datasets.}
\label{tab:real_world_comparison}
\centering
\resizebox{\textwidth}{!}{
\begin{tabular}{l|cc|ccc|cc|cc}
\hline
& \multicolumn{2}{c|}{\textbf{Global Rotation Initialization (deg)}}
& \multicolumn{3}{c|}{\textbf{Fixed-Rotation Global Translation (scene units)}}
& \multicolumn{4}{c}{\textbf{Complete Pipelines}} \\
\cline{7-10}
\multirow{2}{*}{\textbf{Scene (Cams)}}
& \multirow{2}{*}{\textbf{\GLOMAPAvg}\cite{glomap}} & \multirow{2}{*}{\textbf{GRAM}}
& \multirow{2}{*}{\textbf{GLOMAP-GP}\cite{glomap}} & \multirow{2}{*}{\textbf{LiGT}\cite{QiCai_TPAMI}} & \multirow{2}{*}{\textbf{ART}}
& \multicolumn{2}{c|}{\textbf{Rotation Error (deg)}}
& \multicolumn{2}{c}{\textbf{Global-Translation Error (scene units)}} \\
& & & & &
& \textbf{GLOMAP}\cite{glomap} & \textbf{Our Pipeline}
& \textbf{GLOMAP}\cite{glomap} & \textbf{Our Pipeline} \\
\hline
Herz-Jesus-P25 (25)
& \secondresult{0.0397} & \bestresult{0.0327}
& \secondresult{0.0105} & 0.0445 & \bestresult{0.0092}
& \secondresult{0.0284} & \bestresult{0.0219}
& \secondresult{0.0055} & \bestresult{0.0054} \\
Herz-Jesus-P8 (8)
& \secondresult{0.0207} & \bestresult{0.0115}
& 0.0055 & \secondresult{0.0050} & \bestresult{0.0049}
& \secondresult{0.0151} & \bestresult{0.0110}
& \secondresult{0.0033} & \bestresult{0.0025} \\
Castle-P19 (19)
& \secondresult{0.0920} & \bestresult{0.0831}
& \secondresult{0.7017} & 5.5522 & \bestresult{0.0532}
& \bestresult{0.0341} & \secondresult{0.0725}
& \bestresult{0.0196} & \secondresult{0.0469} \\
Castle-P30 (30)
& \secondresult{0.1022} & \bestresult{0.1008}
& \secondresult{0.3553} & 3.9241 & \bestresult{0.0402}
& \secondresult{0.0410} & \bestresult{0.0343}
& \secondresult{0.0204} & \bestresult{0.0188} \\
Entry-P10 (10)
& \secondresult{0.0269} & \bestresult{0.0154}
& \secondresult{0.0325} & 0.0855 & \bestresult{0.0083}
& \secondresult{0.0222} & \bestresult{0.0141}
& \secondresult{0.0052} & \bestresult{0.0050} \\
Fountain-P11 (11)
& \secondresult{0.0277} & \bestresult{0.0143}
& \secondresult{0.0055} & \bestresult{0.0051} & 0.0057
& \secondresult{0.0273} & \bestresult{0.0143}
& \secondresult{0.0023} & \bestresult{0.0021} \\
\hline
courtyard (38)
& \secondresult{0.1956} & \bestresult{0.1068}
& \bestresult{0.0320} & 4.2683 & \secondresult{0.0355}
& \bestresult{0.0218} & \secondresult{0.0270}
& \bestresult{0.0035} & \secondresult{0.0089} \\
delivery\_area (44)
& \secondresult{0.0545} & \bestresult{0.0188}
& \secondresult{0.0231} & 0.8380 & \bestresult{0.0104}
& \bestresult{0.0167} & \secondresult{0.0176}
& \bestresult{0.0025} & \secondresult{0.0045} \\
electro (45)
& \secondresult{0.0728} & \bestresult{0.0325}
& \secondresult{0.5064} & 2.4214 & \bestresult{0.1676}
& \secondresult{0.1161} & \bestresult{0.0184}
& \secondresult{0.0405} & \bestresult{0.0135} \\
facade (76)
& \secondresult{0.0596} & \bestresult{0.0220}
& \secondresult{0.0263} & 8.5180 & \bestresult{0.0183}
& \secondresult{0.0334} & \bestresult{0.0188}
& \secondresult{0.0119} & \bestresult{0.0056} \\
kicker (31)
& \secondresult{0.0716} & \bestresult{0.0381}
& \secondresult{0.0329} & 1.7055 & \bestresult{0.0227}
& \secondresult{0.0172} & \bestresult{0.0128}
& \secondresult{0.1210} & \bestresult{0.0023} \\
meadow (15)
& \secondresult{0.3308} & \bestresult{0.1328}
& \secondresult{0.3490} & 2.8376 & \bestresult{0.1468}
& \secondresult{0.0502} & \bestresult{0.0386}
& \secondresult{0.0191} & \bestresult{0.0178} \\
office (26)
& \secondresult{0.1526} & \bestresult{0.0502}
& \secondresult{0.0169} & 0.4337 & \bestresult{0.0076}
& \secondresult{0.0332} & \bestresult{0.0332}
& \secondresult{0.0019} & \bestresult{0.0016} \\
pipes (14)
& \secondresult{0.1044} & \bestresult{0.0451}
& \secondresult{0.0102} & 0.0834 & \bestresult{0.0056}
& \secondresult{0.0187} & \bestresult{0.0170}
& \secondresult{0.0010} & \bestresult{0.0009} \\
playground (38)
& \secondresult{0.0523} & \bestresult{0.0184}
& \secondresult{0.0189} & 0.0730 & \bestresult{0.0078}
& \secondresult{0.0220} & \bestresult{0.0188}
& \secondresult{0.0015} & \bestresult{0.0014} \\
relief (31)
& \secondresult{0.0290} & \bestresult{0.0183}
& \secondresult{0.0552} & 0.1755 & \bestresult{0.0015}
& \secondresult{0.0278} & \bestresult{0.0132}
& \secondresult{0.0011} & \bestresult{0.0009} \\
relief\_2 (31)
& \secondresult{0.0265} & \bestresult{0.0136}
& \secondresult{0.0052} & 0.3572 & \bestresult{0.0019}
& \secondresult{0.0189} & \bestresult{0.0125}
& \secondresult{0.0010} & \bestresult{0.0008} \\
terrace (23)
& \secondresult{0.0254} & \bestresult{0.0214}
& \bestresult{0.0036} & 0.0082 & \secondresult{0.0057}
& \bestresult{0.0166} & \secondresult{0.0170}
& \bestresult{0.0017} & \secondresult{0.0042} \\
terrains (42)
& \secondresult{0.1746} & \bestresult{0.1086}
& \secondresult{0.0083} & 0.1337 & \bestresult{0.0071}
& \secondresult{0.0284} & \bestresult{0.0254}
& \secondresult{0.0011} & \bestresult{0.0011} \\
\hline
\textbf{Best / 19}
& 0 & \textbf{19}
& \textbf{2} & 1 & \textbf{16}
& 4 & \textbf{15}
& 4 & \textbf{15} \\
\hline
\end{tabular}
}
\raggedright
\vspace{1mm}
\footnotesize{\textit{Note:} Rankings use unrounded values. Red and black boldface indicate the best and second-best results.}
\end{table*}
\begin{table}[!t]
\caption{Median  pipeline runtime over real-world scenes. }
\label{tab:pipeline_runtime}
\centering
{\footnotesize
\setlength{\tabcolsep}{3pt}
\renewcommand{\arraystretch}{1.08}
\begin{tabularx}{\columnwidth}{@{}>{\raggedright\arraybackslash}Xr>{\raggedright\arraybackslash}Xr@{}}
\toprule
\multicolumn{2}{c}{\textbf{GLOMAP}} & \multicolumn{2}{c}{\textbf{Ours}} \\
\cmidrule(lr){1-2}\cmidrule(lr){3-4}
\textbf{Stage} & \textbf{Time}(s) & \textbf{Stage} & \textbf{Time}(s) \\
Sampson two-view refinement & 8.330 & GRAM & 0.042 \\
Rotation Avg. & 0.002 & MORE & 0.621 \\
GLOMAP-GP & 7.526 & ART & 0.043 \\
Triangulation & 0.935 & & \\
BA & 2.850 & & \\
\textbf{Total} & 16.781 & \textbf{Total} & \textbf{0.809} \\
\bottomrule
\end{tabularx}
}
\end{table}

\subsubsection{Global Rotation Initialization}
We compare \GLOMAPAvg{} and GRAM across the four synthetic settings. \GLOMAPAvg{} applies GLOMAP rotation averaging to the relative rotations refined by PoseLib Sampson. As shown in Fig.~\ref{fig:global_simulation_two_method_comparison}\subref{fig:verified_global_rotation_compact}, GRAM achieves lower median errors across all four settings. In the first three settings, the reduction is typically around $20\%$--$30\%$. Under holoplane degeneracy, PoseLib Sampson becomes less reliable, and \GLOMAPAvg{} exhibits a pronounced upper tail. In contrast, component-wise consensus enables GRAM to remain stable with a compact error distribution. GRAM reduces the median error by about half or more at low noise and by roughly $20\%$--$30\%$ at higher noise levels. Further comparisons~\cite{ChatterjeePAMI2018,ErikssonStrongDuality2018,DellaertGTSAM2012,DellaertShonan2020,Martinec07,ChenHybrid2021} show the same overall advantage over additional rotation-averaging methods.

\subsubsection{Global Translation with \GLOMAPAvg{} Rotations}
Following the GLOMAP pipeline, GLOMAP-GP, LiGT, and ART use the same \GLOMAPAvg{} rotations as fixed inputs. Figure~\ref{fig:global_simulation_two_method_comparison}\subref{fig:verified_translation_compact} shows that ART achieves the lowest median error among the three methods across all four settings and noise levels. In the first three settings, ART consistently gives the lowest median error, with a reduction of about 30\%--40\% for $\sigma \leq 2$ pixels compared with GLOMAP-GP and LiGT. As the image noise increases, the three methods become closer. Under holoplane degeneracy, all methods show larger median errors and wider error ranges. GLOMAP-GP exhibits a pronounced upper tail, whereas ART remains stable by jointly aligning LiGT-based triplet translations and retains the lowest median error. Further comparisons~\cite{LUD,1dsfm,BATA2018,OpenMVGPierre,HETA2024} show the same overall trend against additional baselines.

\subsubsection{Global Rotation in Complete Pipelines}
Conventional GLOMAP uses Nist\'er five-point RANSAC, $\ell_2$ Sampson refinement, rotation averaging, GLOMAP-GP, native COLMAP reconstruction, and fixed-intrinsic global BA. Figure~\ref{fig:global_simulation_two_method_comparison}\subref{fig:verified_full_rotation_compact} shows comparable accuracy between the pipelines for $\sigma\leq2$ pixels in the first three settings. At higher noise levels, GRAM+MORE increasingly outperforms GLOMAP, reducing the median error by roughly $50\%$--$75\%$ at $\sigma=5$--$10$ pixels. Under holoplane degeneracy, errors accumulate through the conventional GLOMAP pipeline, leading to a pronounced upper tail. Our pipeline suppresses error propagation through robust rotation initialization and subsequent refinement, maintaining a compact error distribution. Our pipeline shows a more pronounced advantage in median error as noise increases, with reductions of about $60\%$--$80\%$ at $\sigma=5$--$10$ pixels.

\subsubsection{Global Translation in Complete Pipelines}

As shown in Fig.~\ref{fig:global_simulation_two_method_comparison}\subref{fig:verified_art_after_more_center}, the two pipelines are comparable for $\sigma\leq2$ pixels in the first three settings. From $\sigma=4$ pixels onward, our pipeline reduces the median error by roughly $40\%$ or more. Under holoplane degeneracy, GLOMAP exhibits a pronounced upper tail after BA. Our pipeline maintains a narrow error distribution owing to robust rotation initialization and translation estimation. The advantage becomes more pronounced as noise increases, with median-error reductions of about $50\%$--$75\%$ at $\sigma=5$--$10$ pixels.

\subsection{Real-World Experiments}

\subsubsection{Global Rotation Initialization}
\GLOMAPAvg{} and GRAM follow the same rotation-initialization configurations as in the synthetic experiments. Table~\ref{tab:real_world_comparison} shows lower GRAM median rotation error on all 19 scenes, with a median reduction of $47\%$ and the largest  reductions of about $65\%$ on office, delivery\_area, and playground, respectively. Reliable components and cross-component consensus limit unstable two-view estimates. Further comparisons~\cite{ChatterjeePAMI2018,ErikssonStrongDuality2018,DellaertGTSAM2012,DellaertShonan2020,Martinec07,ChenHybrid2021} appear in the supplementary material.

\subsubsection{Global Translation with Fixed Rotations}
GLOMAP-GP, LiGT, and ART share GLOMAP's two-view stage and rotation-averaging result. ART achieves the lowest median global-translation error on 16/19 scenes (Table~\ref{tab:real_world_comparison}). Across the 19 scenes, the median reduction over GLOMAP-GP is $55\%$, with the largest reductions on relief ($97\%$), Castle-P19 ($92\%$), Castle-P30 ($89\%$), and Entry-P10 ($74\%$). 
ART robustly recovers global translations by jointly aligning LiGT-based triplet translations.
Further comparisons~\cite{LUD,1dsfm,BATA2018,OpenMVGPierre,HETA2024} appear in the supplementary material.

\subsubsection{Global Rotation and Translation in Complete Pipelines}

Our pipeline recovers all cameras in all 19 scenes and yields lower rotation and translation errors than GLOMAP on 15/19 scenes each (Table~\ref{tab:real_world_comparison}). MORE lowers GRAM's rotation error on 17/19 scenes, with a median scene-wise reduction of $28\%$ and the largest reductions on terrains ($77\%$), courtyard ($75\%$), kicker ($66\%$), and pipes ($62\%$). Relative to GLOMAP, our pipeline reduces rotation and translation errors on electro by $84\%$ and $67\%$, respectively, and translation error on kicker by $98\%$. Further comparisons with complete SfM systems~\cite{COLMAP2016,OpenMVGPierre,MGSfM2025,FastMap2025} are reported in the supplement.

\subsubsection{Runtime}
Table~\ref{tab:pipeline_runtime} compares the reported runtimes after the shared five-point RANSAC stage; this common stage is excluded from both totals. GLOMAP includes triangulation and fixed-intrinsic BA, whereas our pipeline ends after ART. Using the median of the corresponding per-scene runtime ratios, our pipeline is about $16\times$ faster overall, while GRAM and ART are each more than $100\times$ faster than Sampson refinement and GLOMAP-GP, respectively. Detailed per-scene runtimes are provided in the supplementary material.

\section{Conclusion}

Extending rotation-only imaging geometry, GRAM and MORE employ optimization formulations for global rotation initialization and refinement, respectively, whereas ART utilizes analytical geometric solving to recover global translations. Synthetic and real-world experiments demonstrate complete, accurate, and robust global pose recovery under image noise, corrupted feature matches, local pure rotation, and planar degeneracy. Compared with conventional methods, our approach provides a substantial speed advantage as well. More broadly, our framework might provide a principled basis for streamlining redundant stages in conventional SfM pipelines and restructuring global pose estimation around geometry-driven optimization.

{
\small
\bibliographystyle{ieeenat_fullname}
\bibliography{main}
}

\end{document}